\documentclass[tablecaption=bottom]{jmlr}% journal article

\usepackage{booktabs}
\usepackage{siunitx}
\usepackage{multicol}
\usepackage{multirow}
\usepackage{caption}

\usepackage{adjustbox}
\usepackage{tcolorbox}
\usepackage{booktabs}
\usepackage{multirow}

\usepackage{listings}
\usepackage{xcolor}
\usepackage{adjustbox}
\usepackage{amsmath}  % For general mathematical formatting
\usepackage{amssymb}  % For additional symbols, if needed
\usepackage[T1]{fontenc}  % Ensures proper encoding for texttt

\usepackage[switch]{lineno}

\theorembodyfont{\upshape}
\theoremheaderfont{\scshape}
\theorempostheader{:}
\theoremsep{\newline}

\jmlrvolume{LEAVE UNSET}
\jmlryear{2025}
\jmlrsubmitted{LEAVE UNSET}
\jmlrpublished{LEAVE UNSET}
\jmlrworkshop{Conference on Health, Inference, and Learning (CHIL) 2025} % W&CP title

\title[Synethic EHR via LLMs]{A Case Study Exploring the Current Landscape of Synthetic Medical Record Generation with Commercial LLMs}

\author{%
\Name{Yihan Lin} \Email{yihan23@g.ucla.edu}\\
\addr Department of Computer Science\\ 
University of California, Los Angeles, USA
\AND
\Name{Zhirong Yu} \Email{bellabruin5711@g.ucla.edu}\\
\addr  Bioinformatics IDP\\University of California, Los Angeles, USA
\AND
\Name{Simon A. Lee} \Email{simonlee711@g.ucla.edu}\\
\addr  Department of Computational Medicine\\University of California, Los Angeles, USA
}

\begin{document}

\maketitle

\begin{abstract}
Synthetic Electronic Health Records (EHRs) offer a valuable opportunity to create privacy-preserving and harmonized structured data, supporting numerous applications in healthcare. Key benefits of synthetic data include precise control over the data schema, improved fairness and representation of patient populations, and the ability to share datasets without concerns about compromising real individuals' privacy. Consequently, the AI community has increasingly turned to Large Language Models (LLMs) to generate synthetic data across various domains. However, a significant challenge in healthcare is ensuring that synthetic health records reliably generalize across different hospitals, a long-standing issue in the field. In this work, we evaluate the current state of commercial LLMs for generating synthetic data and investigate multiple aspects of the generation process to identify areas where these models excel and where they fall short. Our main finding from this work is that while LLMs can reliably generate synthetic health records for smaller subsets of features, they struggle to preserve realistic distributions and correlations as the dimensionality of the data increases, ultimately limiting their ability to generalize across diverse hospital settings.
\end{abstract}

\paragraph*{Data and Code Availability}
This work was conducted using numerous enterprise accounts of various commercial Large Language Models. Model Checkpoints may affect reproducibility of this work. The validation data was sourced from eICU database \citep{pollard2018eicu} which is a multi-center dataset comprising deidentified health data from over 200,000 ICU admissions across the United States between 2014 and 2015.

\paragraph*{Institutional Review Board (IRB)}
Our work did not require IRB approval.

\begin{figure}[t!]
\begin{tcolorbox}[colframe=black!75, colback=white, title=Overview of Our Method]
\centering
\includegraphics[width=0.7\textwidth]{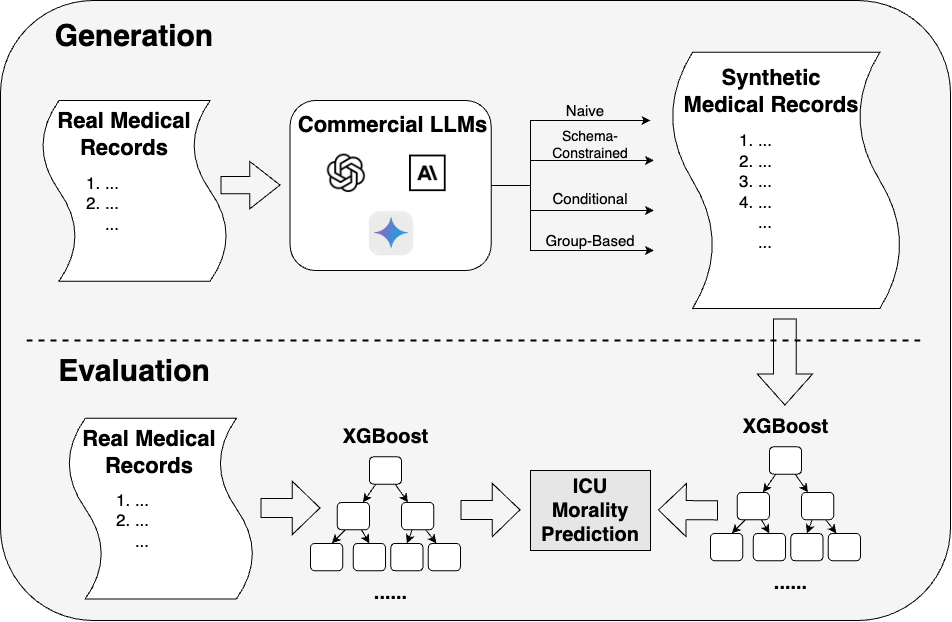}
\captionof{figure}{Overview of our Method}
\label{fig:overview}
\end{tcolorbox}
\end{figure}

\section{Introduction}
\label{sec:intro}

Large Language Models (LLMs) have significantly advanced AI research, serving as powerful tools for diverse applications through their sophisticated natural language understanding and generation capabilities \citep{wang2023improving}. A prominent application of LLMs is synthetic data generation, particularly in healthcare, where they can create synthetic electronic health records (EHRs) \citep{chen2021synthetic}. Synthetic EHRs offer structured and consistent data generation while addressing privacy concerns associated with real patient data. Additionally, LLMs can generate datasets that better represent underrepresented and marginalized groups, mitigating the diversity limitations of traditional datasets such as MIMIC \citep{johnson2016mimic} and UK Biobank \citep{bycroft2018uk}.

Despite these advantages, generating synthetic medical records poses challenges in ensuring model generalizability across diverse patient populations and heterogeneous data structures \citep{goetz2024generalization, goldstein2017opportunities}. Efforts in data harmonization, including frameworks like MEDS \citep{ kolo2024meds} and schema-matching techniques \citep{parciak2024schema}, have made strides but have not fully resolved these issues. Consequently, synthetic data generation remains a promising approach for creating harmonized and adaptable datasets.

In this work, we explore the use of LLMs to generate synthetic EHRs and evaluate their effectiveness in healthcare modeling. We investigate various generation strategies, focusing on factors such as sample size, dimensionality, as well as fidelity and privacy. To assess the generalizability of the synthetic data, we conduct multi-site validation using real data from the eICU database \citep{pollard2018eicu}. Our benchmarking framework uses an XGBoost \citep{shwartz2022tabular} to provide a robust evaluation of synthetic datasets, addressing the gap in assessing the robustness and generalizability of LLM-generated synthetic data in healthcare.

\paragraph{Importance of the Problem}

Fair and representative datasets are crucial for developing equitable and effective AI systems in healthcare \citep{chen2023algorithmic}. The field faces persistent challenges related to data access, diversity, and bias, which compromise the reliability and fairness of AI models \citep{chen2018my}. Synthetic data generation offers a viable solution to these issues by providing alternatives that enhance diversity and protect privacy. However, there is a lack of rigorous studies evaluating the robustness and generalizability of synthetic datasets, particularly in healthcare settings. 

\paragraph{What Makes Generating EHR Challenging?}

Generating synthetic EHR presents significant challenges, primarily due to the intricate and clinically meaningful relationships that must be preserved between covariates and features. EHR data encompasses a wide array of variables, each of which interacts in complex, non-linear ways. Ensuring that these interdependencies remain coherent and reflective of real-world medical scenarios is crucial for the synthetic data to be both useful and valid for downstream applications.

As the scale increases, maintaining these intricate relationships becomes exponentially more difficult. High-dimensional data introduces issues such as sparsity and the curse of dimensionality, which complicate the modeling of joint distributions and the preservation of conditional dependencies. Balancing the fidelity of synthetic data with computational feasibility and privacy constraints further exacerbates the difficulty, making the generation of large-scale, realistic EHR datasets a formidable task.

\section{Related Works}
\subsection{Synthetic Data Generation}

Synthetic data—artificially generated information that replicates the statistical properties of real-world data—has become a pivotal resource across various industries \citep{raghunathan2021synthetic, jordon2022synthetic, nikolenko2021synthetic}. Its benefits include enhanced privacy by removing personal identifiers, reduced data collection costs, and the ability to generate large-scale, tailored datasets. Nonetheless, challenges such as the potential omission of rare edge cases and the need for rigorous validation to ensure accuracy and relevance persist.

In the context of Large Language Models (LLMs), synthetic data offers significant research advancements. It enables LLMs to learn from a broader spectrum of examples without exposing sensitive information or infringing on proprietary content \citep{gholami2023does}. Ensuring the quality and relevance of synthetic data is crucial, as inaccuracies can impair model performance. Studies have shown that appropriately generated synthetic data can enhance LLM performance on downstream tasks \citep{gholami2023does}, improve hidden state representations through pre-training \citep{wang2023improving}, and facilitate complex reasoning in applications like AlphaGeometry \citep{trinh2024solving}.

Recent advancements indicate that LLMs surpass traditional generative models, such as Generative Adversarial Networks (GANs) and Variational Autoencoders (VAEs), in producing high-fidelity synthetic tabular data \citep{borisov2023language}. For example, GReaT \citep{borisov2023language} leverages pre-trained LLMs to outperform GANs in synthesizing high-quality tabular data. Subsequent models, including ReaLTAbFormer \citep{solatorio2023real}, TabuLa \citep{zhao2023tabula}, DP-LLMTGen \citep{tran2024differential}, and CLLM \citep{seedat2024curated}, have introduced new features that enhance data generation capabilities. Additionally, MALLM-GAN \citep{ling2024mallmgan} integrates LLMs within GAN architectures to further improve synthetic data generation.

\subsection{Synthetic Data in Healthcare}

Synthetic data is well-established in healthcare, with studies evaluating its benefits and challenges \citep{gonzales2023synthetic}. The high-dimensional nature of patient records requires advanced methods for accurate generation. Generative Adversarial Networks (GANs) have been pivotal in this domain, capable of simulating the complex distributions of patient data \citep{yan2024generating}. Notable models include PATE-GAN \citep{jordon2018pate}, which incorporates differential privacy via the Private Aggregation of Teacher Ensembles (PATE) approach, ADS-GAN \citep{yoon2020anonymization}, TimeGAN \citep{yoon2019time}, and attentive state-space models \citep{NEURIPS2019_1d0932d7}, each enhancing data quality and privacy in different ways. Tools such as GOGGLE \citep{liu2023goggle} and DECAF \citep{vanbreugel2021decaf} focus on generating high-fidelity and fair synthetic tabular data, respectively.

LLMs have also emerged as powerful tools for generating synthetic Electronic Health Records (EHRs), addressing data scarcity and privacy concerns in medical research. By leveraging extensive biomedical literature and medical records, LLMs can produce realistic patient data that mirrors real-world datasets without compromising patient confidentiality \citep{hao2024llmsyn}.

\subsection{Leveraging Large Language Models in Healthcare}

Large Language Models (LLMs) have been increasingly utilized in healthcare to enhance patient care and clinical decision-making. Recent advancements have led to the development of models such as MOTOR \citep{steinberg2023motortimetoeventfoundationmodel} and Event Stream GPT (ESGPT) \citep{mcdermott2023eventstreamgptdata}, which are pre-trained on Electronic Health Record (EHR) data to capture complex event sequences in continuous time. Additionally, approaches such as MEME \citep{lee2024multimodal} and GenHPF \citep{hur2023genhpf} enable the transformation of structured EHR data into textual formats \citep{hegselmann2023tabllm, ono2024text}, facilitating the application of LLMs to various predictive tasks within the language modeling space. The incorporation of inductive biases, as demonstrated by DK-BEHRT \citep{an2025}, and Clinical ModernBERT \citep{lee2025clinical}, along with the integration of external knowledge bases \citep{wang2024large}, further improves the  performance and reliability of LLMs in clinical applications.

Beyond structured data modeling, LLMs have demonstrated significant potential in clinical decision support, including disease diagnosis \citep{zhou2024large}, personalized medication recommendations \citep{lee2024enhancing}, treatment optimization \citep{benary2023leveraging}, automated medical coding \citep{soroush2024large, lee2024large}, and clinical document generation \citep{yuan2024continued, kumichev2024medsyn}. In medical question-answering tasks, models such as Med-PaLM 2 \citep{singhal2023expert} have achieved notable performance improvements, outperforming previous models on benchmarks such as MedQA \citep{jin2021disease} and MedExQA \citep{kim-2024-medexqa}. Recent research has focused on refining training methodologies and incorporating external medical knowledge sources to improve the factual accuracy and contextual relevance of LLM-generated responses \citep{yang2023integrating}, further advancing their potential for real-world deployment in clinical settings.

\section{Methods}

\subsection{Large Language Models \& Data Generation}

In our study, we leverage ChatGPT Enterprise \footnote{\url{https://openai.com/index/introducing-chatgpt-enterprise/}} as our primary framework for operating large language model (LLM). In particular we use o1 models to help us generate synthetic data as it represents one of the state of the art commercial LLMs across a broad range of tasks \citep{jaech2024openai}. Further experimentation is done on two other commercial LLMs and there results can be found in the appendix.

The primary objective of this study is to investigate methods for generating synthetic data that effectively generalizes \textbf{within the distribution} of the eICU database \citep{pollard2018eicu}. Specifically, we define generalization as the ability to produce synthetic data, \( \hat{\mathcal{D}} \), such that the distribution of its features, \( P_{\hat{\mathcal{D}}}(\mathbf{x}) \), closely approximates the true data distribution, \( P_{\mathcal{D}}(\mathbf{x}) \), within the same feature space \( \mathbf{x} \in \mathbb{R}^d \). Mathematically, this is expressed as minimizing the divergence between these distributions 
\(
D(P_{\mathcal{D}} \| P_{\hat{\mathcal{D}}}) \quad 
\), 
where  $D(\cdot \|\cdot)$ denotes a divergence measure (via KL divergence).
To achieve this, we aim to generate synthetic datasets adhering to the schema of the eICU database, ensuring that each column corresponds to a predefined feature or label and each row represents a patient or recorded visit. The features included in the generated data are informed by the attributes presented in the \cite{johnson2018generalizability} study.

\paragraph{Naive Generation}
In the naive generation approach, a large language model (LLM) is simply shown an example of the eICU data and asked to produce synthetic EHR rows based on that single example file. No additional instructions or constraints are provided. This technique can be viewed as the most straightforward way (baseline) of generating synthetic records: the model observes the structure, values, and potential distribution of features in a small sample of real data, then attempts to mimic that distribution in its outputs.

\paragraph{Schema-Constrained Generation}
A more refined method similar to that of \citep{borisov2023language} introduces explicit instructions or constraints that the LLM must follow while generating synthetic data.

By emphasizing relevant domain rules, this approach reduces the risk of producing logically inconsistent entries. However, it demands more preparatory work to encode these constraints in the prompt, and extensive prompt engineering is required to balance realism with data diversity. 

\paragraph{Conditional Generation}
A key limitation of purely schema-constrained approaches is the lack of dynamic conditioning on previously generated features. In conditional generation used by many previous works \citep{borisov2023language,vardhan2024large}, each feature is sampled incrementally, taking into account the values already generated. Formally, let $\mathbf{x} = (x_1, x_2, \ldots, x_N)$ represent the $N$ features (e.g., vital signs, demographic attributes, lab results) for a patient record. The LLM approximates the joint distribution
\begin{equation}
    P(\mathbf{x}) \;=\; P(x_1, x_2, \ldots, x_N)\;=\;\prod_{i=1}^{N} P\bigl(x_i \,\bigm|\, x_1,\ldots,x_{i-1}\bigr).
\end{equation}
In practice, the model sequentially generates $x_i$ conditioned on all previously generated features $(x_1, \dots, x_{i-1})$. For example, if $x_1$ (age) is generated to be $75$, the conditional distribution for $x_2$ (heart rate) can be biased toward geriatric norms. This “chain-of-thought” \citep{wei2022chain} style conditioning allows the model to maintain more realistic dependencies among features and minimize inconsistencies (e.g., contradictory comorbidities).

\paragraph{Group-Based Generation Approach}

The group-based generation approach introduces a demographic subpopulation variable \( g \) to condition the synthetic data generation process. This method allows the model to capture group-specific patterns in the data, ensuring that the generated records reflect the unique distributions observed in different demographic groups.

For example, let \( g \in \{\text{Male}, \text{Female}\} \) represent the group variable encoding gender. In this approach, the model first samples a value for \( g \) (e.g., \( g = \text{Male} \)), and then generates all features \( \mathbf{x} = \{x_1, x_2, \ldots, x_N\} \) conditioned on this group label. Formally, the generation process can be expressed as:
\[
P(\mathbf{x} \mid G=g) = \prod_{i=1}^N P\bigl(x_i \mid x_1, \ldots, x_{i-1}, G=g\bigr).
\]

This conditioning ensures that the synthetic data captures meaningful correlations between demographic factors and clinical attributes, improving the representativeness of the generated dataset. In our study we use race and gender as our group variable \( g \) and ask the LLM to perform a uniform number of samples for all groups.

\begin{table*}[t!]
\begin{tcolorbox}[colframe=black!75, colback=white, title=Performance Comparison of Generative Strategies]
\centering
\begin{adjustbox}{width=1\textwidth}
\begin{tabular}{@{}lcccc@{}}
\toprule
\textbf{Strategy/Features/Sample Size} & \textbf{Within/Across Dataset} & \textbf{Avg. KL Divergence}       & \textbf{AUC (Mean ± CI)}       & \textbf{AUPRC (Mean ± CI)}     \\ 
\midrule
\multirow{2}{*}{Naive/all/1k} & within       & ---  & 0.4558 [0.3772, 0.5451] & 0.4347 [0.3518, 0.5352] \\
 & across       & 0.5797 & 0.5382 [0.4829, 0.5943] & 0.5398 [0.4585, 0.6084] \\
\midrule
\multirow{2}{*}{Schema/all/1k}  & within       & ---  & 0.4319 [0.3579, 0.5259] & 0.5082 [0.4234, 0.5970] \\
  & across       & 0.5212 & 0.6205 [0.5623, 0.6848] & 0.5774 [0.4974, 0.6528] \\
\midrule
\multirow{2}{*}{Conditional/all/1k}  & within       & ---  & 0.5051 [0.4207, 0.5951] & 0.4099 [0.3351, 0.4908] \\
  & across       & 0.3051 & 0.4769 [0.4279, 0.5358] & 0.4858 [0.4230, 0.5608] \\
\midrule
\multirow{2}{*}{Group/all/1k}  & within       & --- & 0.5136 [0.4275, 0.5942] & 0.5341 [0.4352, 0.6326] \\
 & across       & \textbf{0.2963} & 0.5052 [0.4472, 0.5634] & 0.5070 [0.4446, 0.5849] \\
\bottomrule
\end{tabular}
\end{adjustbox}
\caption{Performance comparison of different generative strategies for synthetic data generation across all features and 1,000 samples. Metrics include average KL divergence, AUC (Mean ± Confidence Interval), and AUPRC (Mean ± Confidence Interval), evaluated in both within-dataset and across-dataset scenarios.}
\label{tab-results}
\end{tcolorbox}
\end{table*}

\section{Experimental Setup}
\subsection{Benchmarking and Evaluation}

To validate the robustness and generalizability of our synthetic data generation approach, we established a comprehensive benchmarking framework encompassing critical factors such as sample size and feature dimensionality. We used XGBoost \citep{shwartz2022tabular} as our baseline model due to its proven efficacy in tabular data tasks, aligning with having a singular baselines of assessing generalizability from prior studies \citep{johnson2018generalizability}.

\paragraph{Experimental Setup}
Our experiments involved generating 1,000 synthetic samples, each comprising 83 features, using large language models (LLMs) based on the strategies detailed in Section 3. To ensure consistent evaluation, we maintained constant dataset sizes across different feature subsets, systematically varying the number of features to isolate dimensionality effects. This setup enabled us to benchmark the stability of high-dimensional datasets and the fidelity of the synthetic data produced by various generation strategies.

\paragraph{Prediction Task}
We utilized the eICU Collaborative Research Database \citep{pollard2018eicu} to develop and evaluate models for predicting ICU mortality, a pivotal task in AI for healthcare \citep{arnrichmedical}. This binary classification problem determines whether a patient dies during their ICU stay, leveraging the database's diverse and multi-institutional records to test generalizability.

\paragraph{Model Training and Evaluation}
Predictive models, including the XGBoost classifier, were trained on both real and synthetic datasets. Performance was measured using the Area Under the Receiver Operating Characteristic Curve (AUROC) and Precision-Recall (PR) metrics. We evaluated models in intra-dataset settings (training and testing on the same dataset) and inter-dataset settings (training on synthetic data and testing on real data), providing a robust assessment of the synthetic data's utility and the generation strategies' effectiveness.

\subsection{Hyperparameter Tuning}

Effective hyperparameter tuning is crucial for optimizing large language models (LLMs) to generate high-quality synthetic data. Our analysis focused on two key hyperparameters: feature dimensionality and training sample size. \textbf{We applied our findings, leveraging the optimal strategy identified in each stage of analysis to guide subsequent evaluations.}

\subsubsection{Feature Dimensionality}

To evaluate the impact of feature count, we trained models using datasets containing the top 5, 10, 15, and 20 features, selected based on feature importance rankings (Figure in appendix). Each subset maintained a constant dataset size to isolate the effect of dimensionality. 

\subsubsection{Sample Size}

We also investigated training sample sizes of 1,000, 5,000, and 10,000 records to balance data representation and noise. Our motive here was to test whether increasing the sample size resulted in more diverse representations of data or conversely generated adverse examples that may affect overall downstream performance.

\subsection{KL Divergence as a Measure of Fidelity}

Kullback-Leibler (KL) divergence \citep{kullback1951kullback} serves as a fundamental metric for quantifying the discrepancy between two probability distributions. In this study, we use KL divergence to evaluate the fidelity of synthetic data by comparing the marginal distributions of each feature in the real data (\(P\)) against those in the synthetic data (\(Q\)). We also use it as a metric for fairness comparing the divergences across different demographic and gender groups.  Mathematically, KL divergence is defined as:

\[
D_{\text{KL}}(P \parallel Q) = \int_{\mathcal{X}} P(x) \log \frac{P(x)}{Q(x)} \, \mathrm{d}x
\]

where \(\mathcal{X}\) denotes the set of possible feature values. A lower KL divergence value indicates a closer alignment between the synthetic and real data distributions, thereby reflecting higher fidelity of the synthetic data.

\subsection{Privacy Assessment}
Ensuring the privacy of individuals represented in synthetic datasets is paramount, particularly within the sensitive context of healthcare. To comprehensively evaluate the privacy risks associated with our EHRs, we used an approach on Membership Inference Attacks (MIAs) \citep{hu2022membership} to test its privacy.

MIAs were conducted to determine whether an adversary could accurately identify if a specific data point was part of the original training dataset used to generate the synthetic data. By training an attack model on features derived from model outputs, such as prediction probabilities and dataset characteristics, we assessed the model's ability to distinguish between members and non-members. The effectiveness of these attacks was quantified using the Area Under the Receiver Operating Characteristic Curve (AUROC), Membership Advantage, and Empirical Risk. High AUROC and Membership Advantage values indicate a greater susceptibility to MIAs, whereas minimal differences in Empirical Risk suggest stronger privacy preservation.

\section{Results}

\subsection{Generation Strategies Comparison}

In Table~\ref{tab-results}, we present the results of evaluating XGBoost across various synthetic data generation strategies. Despite generating only 1,000 samples across 83 features, none of the tested strategies consistently demonstrated robust performance across AUC or AUPRC. XGBoost exhibited highly variable outcomes, with modest improvements in one metric often accompanied by declines in another. Within-dataset scenarios tended to yield slightly better results; however, these improvements were trivial and did not generalize effectively to real datasets.

When examining the KL divergence of continuous features for each generation strategy, we observed a consistent decrease in average KL divergence across successive generative methods. This indicates that while certain features are better aligning with the true data distributions, adverse features may still be introducing inconsistencies that hinder the model's ability to predict ICU mortality. Consequently, these misaligned features likely contribute to the failure of the synthetic datasets to generalize effectively.

\begin{table}[t!]
\begin{tcolorbox}[colframe=black!75, colback=white, title=Inspecting fairness in generation]
\centering
\begin{adjustbox}{width=0.5\columnwidth}
\begin{tabular}{lcc}
\hline
\textbf{Demographic Column} & \textbf{Avg. KLD (Group)} & \textbf{Avg. KLD (Naive)} \\ \hline
race\_black   & \textbf{0.329642}  & 0.547470                      \\
race\_hispanic  & \textbf{0.362365}  & 0.484697                     \\
race\_asian   & \textbf{0.334025 } & 0.566793                    \\
race\_other   & \textbf{0.399359}  & 0.440829                     \\
is\_female   & \textbf{0.393389}  & 0.411177                    \\ \hline
\end{tabular}
\end{adjustbox}
\caption{Average KL Divergence per Demographic Group}
\label{tab:kl_divergence_group1}
\end{tcolorbox}
\end{table}

\subsubsection{A closer look at fairness}

We identified in the preliminary results that the group-based generation strategies yielded a lower KL divergence than all other strategies. However we wanted to inspect in particular whether different demographic groups had a balanced KL divergence or if there were any exacerbated divergences. Table~\ref{tab:kl_divergence_group1} presents the Average KL Divergence (KLD) values for different demographic groups, comparing the results from a group-based generation strategy versus a naive approach. Across all demographics, the group-based strategy generally yields lower KLD values compared to the naive approach. This suggests a more tailored alignment with the underlying data when conditioning on the group variable. While variations in divergence are observed across different demographic categories, the results provide a comparative view of the divergence levels for both strategies without indicating any immediate extreme disparities.

\subsection{Number of Features Experiment}
\begin{table}[h!]
% Table 2: Performance by Number of Features
\begin{tcolorbox}[colframe=black!75, colback=white, title=Performance by Number of Features]
\centering
\begin{adjustbox}{width=0.5\columnwidth}
\begin{tabular}{@{}lccc@{}}
\toprule
\textbf{Number of Features} & \textbf{Scenario} & \textbf{Avg. KL Divergence} & \textbf{AUC (Mean ± Range)} \\ 
\midrule
\multirow{2}{*}{5}   & within  & --- & 0.7791 [0.5366, 0.9892] \\
                     & across  & 0.1690 & 0.8224 [0.8146, 0.8302] \\ 
\midrule
\multirow{2}{*}{10}  & within  & --- & \textbf{0.8710} [0.5635, 1.0000] \\
                     & across  & \textbf{0.1552} & \textbf{0.8133} [0.8038, 0.8229] \\ 
\midrule
\multirow{2}{*}{15}  & within  & --- & 0.4924 [0.4107, 0.5876] \\
                     & across  & 0.2293 & 0.6885 [0.6692, 0.7050] \\ 
\midrule
\multirow{2}{*}{20}  & within  & --- & 0.4833 [0.3953, 0.5542] \\
                     & across  & 0.2466 & 0.5335 [0.5157, 0.5508] \\ 
\bottomrule
\end{tabular}
\end{adjustbox}
\caption{Performance evaluation of synthetic data generation based on number of features. We observe better downstream performance on smaller subset of features.}
\label{tab:auc_results}

\end{tcolorbox}
\end{table}

% Table 3: Performance by Sample Size
\begin{table*}[t!]
\begin{tcolorbox}[colframe=black!75, colback=white, title=Performance by Sample Size (10 Features)]
\centering
\begin{adjustbox}{width=0.7\textwidth}
\begin{tabular}{@{}lcccc@{}}
\toprule
\textbf{Features/Sample Size}               & \textbf{Within/Across Dataset} & \textbf{Avg. KL Divergence} & \textbf{AUC (Mean ± Range)}       & \textbf{AUPRC (Mean ± Range)}          \\ 
\midrule
\multirow{2}{*}{10 features/1k}   & within            & ---  & 0.8710 [0.5635, 1.0000]          & 0.4444 [0.4444, 1.0000]                         \\
                                  & across            & 0.1552 & 0.8133 [0.8038, 0.8229]          & 0.5809 [0.5608, 0.6012]                   \\ 
\midrule
\multirow{2}{*}{10 features/5k}    & within            & ---  & 0.8780 [0.7781, 0.9491]          & 0.7957 [0.6494, 0.9063]                 \\
                                  & across            & 0.0952 & 0.9015 [0.8945, 0.9090]          & 0.7586 [0.7418, 0.7761]                \\ 
\midrule
\multirow{2}{*}{10 features/10k}   & within            & ---  & \textbf{0.9437} [0.9115, 0.9743]          & \textbf{0.8093} [0.6776, 0.9190]              \\
                                  & across            & \textbf{0.0821} & \textbf{0.9157} [0.9089, 0.9217]          & \textbf{0.7969} [0.7797, 0.8122]              \\ 
\bottomrule
\end{tabular}
\end{adjustbox}
\caption{Performance evaluation of synthetic data generation using 10 features with varying sample sizes (1k, 5k, and 10k). Metrics include average KL divergence, AUC (Mean ± Range), and AUPRC (Mean ± Range), assessed in both within-dataset and across-dataset scenarios.}
\label{samplesize}
\end{tcolorbox}
\end{table*}

The next results are a continuation of our findings from Table~\ref{tab-results}, and we continue our analysis by using the group generation strategy as it yielded the best results. Therefore, in  Table~\ref{tab:auc_results}, we highlight the challenges of generating high-dimensional features with an LLM and reveal a clear relationship between the number of features and model performance, as measured by AUC, in both within-distribution and across-distribution scenarios. Notably, the average KL divergence is lower when fewer features are generated, but excessively small feature subsets exhibit greater differences from the real data. This suggests a balance must be struck between too few and too many features for optimal performance.

As the number of features increases beyond 10, model performance degrades significantly. For subsets with 15 or 20 features, AUC values decline, indicating that the added dimensionality over-complicates the generation process, thereby reducing the model's ability to generalize. This trend is further corroborated by average KL divergence scores, which begin to rise again as the feature dimensionality increases, reflecting poorer alignment with the real data distributions.

\subsection{Sample size experiment}

Lastly building off of these experiments we proceed again with the group based generation this time using only 10 features. In our sample size experiment, we examine Table~\ref{samplesize}, where it reveals a consistent trend across different sample sizes and evaluation scenarios. As the sample size increases from 1,000 to 10,000, both within-dataset and across-dataset performance metrics (AUC and AUPRC) improve. This trend suggests that larger sample sizes enable better alignment of the synthetic data with the real data distribution, leading to enhanced model generalizability. However like the dimensionality experiments, we hypothesize that as the models are asked to scale the number of samples generated, they will probably begin to generate adverse or repeating samples.

Additionally, the average KL divergence decreases as the sample size grows, particularly in the across-dataset scenario. This indicates that larger sample sizes result in synthetic data that better approximates the real data's statistical properties, reducing discrepancies between the distributions.

\subsection{Inspecting Distributions of Generated Data}

\begin{table}[t!]
\begin{tcolorbox}[colframe=black!75, colback=white, title=Top 5 and Bottom 5 generated features]
\centering
\begin{adjustbox}{width=0.5\columnwidth}
\begin{tabular}{l|c}
\hline
\textbf{Feature} & \textbf{KL Divergence} \\ \hline
\multicolumn{2}{c}{\textbf{Top 5 Features}} \\ \hline
hosp\_los                & 0.003344 \\ 
is\_female               & 0.002651 \\ 
hemoglobin\_first\_early & 0.013133 \\ 
hematocrit\_last\_early  & 0.016143 \\ 
albumin\_first\_early  & 0.016901 \\ \hline
\multicolumn{2}{c}{\textbf{Bottom 5 Features}} \\  \hline
bilirubin\_last\_early   & 0.835301 \\ 
bilirubin\_first\_early  & 0.822738 \\ 
inr\_last\_early         & 0.675728 \\ 
creatinine\_first\_early & 0.657346 \\ 
creatinine\_last\_early  & 0.632238 \\  \hline
\end{tabular}
\end{adjustbox}
\caption{KL Divergence Scores: Top 5 and Bottom 5 important Features when asked to generate synthetic data for all 83 features}
\label{tab:kl_divergence_top_bottom}
\end{tcolorbox}
\end{table}

In addition to our benchmarking results, we also take a closer look at the KL divergence scores, which reveals insights into the effectiveness of synthetic data generation across different scenarios. When tasked with generating synthetic data for a small subset of features, the model demonstrates strong performance, as reflected by a lower average KL divergence scores for all features. This suggests that the model can accurately replicate the statistical properties of the real data in a constrained setting. 

However, the performance significantly deteriorates when the model is tasked with generating synthetic data for all 83 features (Table~\ref{tab:kl_divergence_top_bottom}). In this table, we take a look at the KL divergences of all the features and find that while some features exhibit low KL divergence and align closely with the real data distribution, others deviate substantially, which contributes to the degradation of predicting mortality. These adverse features introduce inconsistencies/noise into training, which likely affect the downstream performance of models trained on this data.
\vspace{-0.16in}

\subsection{Membership Inference Attacks}

\begin{tcolorbox}[colframe=black!75, colback=white, title=Membership Inference Attack Results]
\centering
\includegraphics[width=0.5\textwidth]{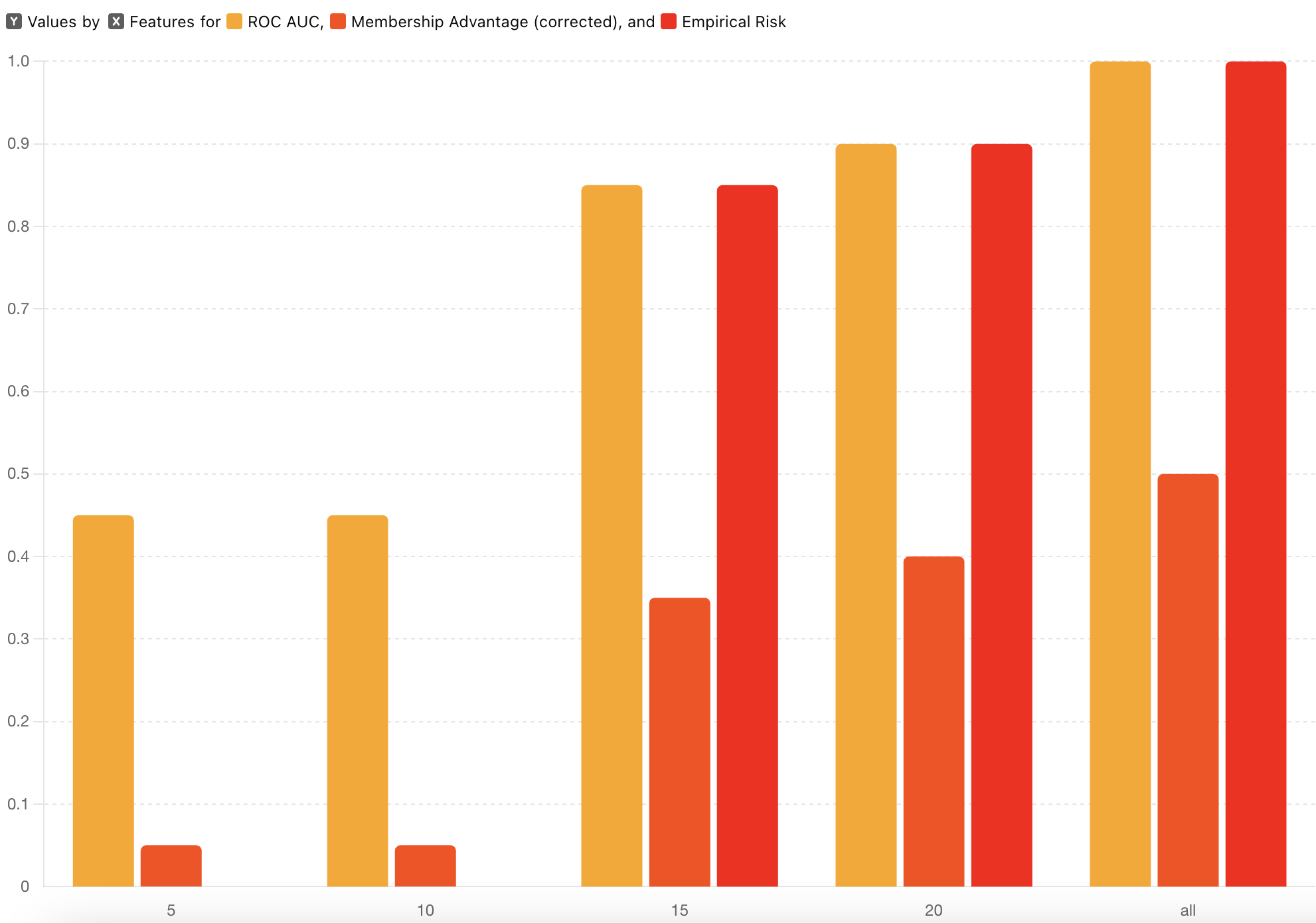}
\captionof{figure}{Bar plot illustrating the Membership Inference Attack Results, comparing AUC, Membership Advantage, and Empirical Risk across different numbers of features.}
\label{fig:mi}
\end{tcolorbox}

The results presented in Figure~\ref{fig:mi} demonstrate the varying effectiveness of membership inference attacks as the number of features increases. For datasets with 5 and 10 features, the attack shows limited effectiveness, with low AUC values (0.4509 and 0.4415) and near-zero membership advantage (0.0011 and 0.0032). These results suggest that the model's outputs for members and non-members are nearly indistinguishable, indicating stronger privacy preservation.

However, for datasets with larger set of features, the attack begin to incrementally increase suggesting that the data becomes less realistic when scaled to higher dimensions. This indicates a complete compromise of privacy, as the attack model can reliably distinguish members from non-members. The empirical risk for these cases is also extremely high, highlighting the significant privacy risks associated with datasets containing a higher number of features.

\section{Discussion}
\subsection{The Challenges of Generating High-Dimensional Data} 
Our experiments highlight the substantial difficulties encountered when generating high-dimensional synthetic data like Electronic Health Records with LLMs. As demonstrated in Table~\ref{tab:auc_results}, increasing the dimensionality from 10 to 15 and 20 features sharply reduces model performance for both within-dataset and across-dataset scenarios. These results underscore two crucial issues. First, models must capture a growing number of complex dependencies among features, an inherently difficult task that often leads to compounding errors and higher divergence from the real data distribution. Second, as dimensionality grows, the generative process becomes increasingly susceptible to overfitting certain feature correlations while failing to capture others. This discrepancy directly impacts the downstream performance of classifiers trained on synthetic data, as observed by the diminishing Area Under the Curve (AUC) scores.

Notably, our results suggest a “sweet spot” around ten features, where the generative model can maintain relatively low divergence and produce synthetic data that yields reasonably strong classifier performance. Beyond this range, the added complexity appears to overwhelm the model, causing degradation in predictive performance. 

We also observed a similar trend in regards to minimizing the divergence between synthetic datasets and ensuring privacy among these different number of features. We found as LLMs were tasked to generate more features, this resulted in a larger average KL divergence and ``less realistic'' and less privacy adhering data. These observations underscore the inherent trade-off between capturing the full richness of a dataset and preserving enough fidelity/privacy to support effective downstream learning tasks.

\subsection{A comment on fairness}

From the results presented in Table \ref{tab:kl_divergence_group1}, it is evident that there is minimal disparity in KL divergence across different racial groups for both the group-based and naive generation strategies. This suggests that neither approach introduces significant inequity in terms of how well the generated data aligns with the underlying distributions of different demographic groups. 

However, a clear pattern emerges when comparing the two strategies: the group-based approach consistently yields lower KL divergence values across all demographic groups. This indicates that conditioning on the group allows for more precise modeling of the underlying data distribution, leading to improved generation performance. The improvement is particularly meaningful as it is achieved without introducing substantial discrepancies between demographic groups, highlighting the potential of group-based strategies to enhance fairness and accuracy simultaneously.

\subsection{KL Divergence and the Impact of Sample Size and Dimensionality on Fidelity}

A thorough examination of KL divergence elucidates how both sample size and dimensionality influence the model's ability to accurately replicate data distributions. When generating a small subset of ten features with a substantial sample size, the average KL divergence remains notably low. This low divergence signifies a strong alignment between the synthetic and real feature distributions, thereby enhancing the fidelity of the generated data. The accompanying improvements in both AUC and AUPRC metrics for these lower-dimensional, adequately sampled scenarios underscore the critical role that distributional fidelity plays in producing reliable synthetic datasets.

Conversely, as dimensionality increases, the fidelity of the synthetic data generation process becomes more susceptible to challenges. When the model attempts to generate all 83 features, even with a large sample size, the average KL divergence escalates. While certain features maintain low KL divergence, a significant number of other features exhibit substantially higher scores. This increase indicates that the generative model struggles to capture essential distributional characteristics across the full feature set. High-dimensional settings exacerbate the difficulty of maintaining fidelity, as the complexity of inter-feature relationships grows. These "adverse" features can compromise classifier performance by introducing misleading patterns and misaligned samples, ultimately resulting in decreased AUC and AUPRC. These observations highlight the necessity of balancing sample size and dimensionality and emphasize the importance of identifying and addressing problematic features to maintain high fidelity in synthetic data generation within high-dimensional spaces.

\subsection{Privacy Implications of Sample Size and Dimensionality in Synthetic Data}

Figure~\ref{fig:mi} presents a comprehensive analysis of how both sample size and dimensionality affect the privacy of synthetic datasets, particularly in the context of membership inference attacks. For datasets with a modest number of features (e.g., 5 and 10 features) and sufficiently large sample sizes, the attack's success is limited, as indicated by low ROC AUC values (0.4509 and 0.4415) and minimal membership advantages (0.0011 and 0.0032). These results suggest that with ample data and lower dimensionality, the model effectively obscures the distinctions between members and non-members, thereby ensuring robust privacy preservation. The low empirical risk in these scenarios further supports the notion that the model generalizes well while safeguarding individual privacy.

However, as dimensionality increases, even with larger sample sizes, the effectiveness of membership inference attacks rises significantly. Higher-dimensional datasets exhibit greater ROC AUC and membership advantage metrics, indicating a more pronounced ability for adversaries to differentiate between members and non-members. This trend is partly attributable to the model's diminished capacity to generalize in high-dimensional spaces, where the complexity of the data can lead to overfitting and leakage of sensitive information. Additionally, larger sample sizes in high-dimensional settings may not proportionally mitigate privacy risks, as the curse of dimensionality can still expose subtle patterns that facilitate membership inference.

Moreover, there is a discernible correlation between membership inference metrics and overall data quality. While higher-dimensional datasets can offer richer and more nuanced representations, they simultaneously present increased privacy challenges. The enhanced detail in such datasets may inadvertently provide adversaries with more vectors to exploit, thereby weakening privacy guarantees. This delicate interplay between sample size, dimensionality, data quality, and privacy underscores the necessity for careful model and dataset design. Particularly in applications where privacy is paramount, it is essential to consider how scaling dimensionality and adjusting sample sizes can impact both the fidelity of synthetic data and its vulnerability to privacy breaches.

\subsection{The Limitations of LLMs in Generating Synthetic Data} 
In this study, we conducted a comprehensive evaluation of Large Language Models (LLMs) in generating synthetic Electronic Health Records (EHRs), with a particular focus on how sample size and feature dimensionality influence both the fidelity, fairness and privacy of the synthetic data. Our experiments revealed that LLMs are capable of producing high-fidelity synthetic data when the number of features is limited. Specifically, subsets containing up to ten features exhibited low Kullback-Leibler (KL) divergence, indicating a strong alignment between the synthetic and real data distributions. This high fidelity was further supported by improved Area Under the Receiver Operating Characteristic Curve (AUC) and Area Under the Precision-Recall Curve (AUPRC) metrics, demonstrating that synthetic data in lower-dimensional settings can effectively support downstream predictive tasks.

However, as the dimensionality of the data increased to encompass all 83 features, the fidelity of the synthetic data generation process significantly declined. The average KL divergence rose substantially, and many features exhibited high divergence scores, highlighting the LLMs' struggle to accurately capture complex inter-feature relationships inherent in high-dimensional EHRs. This deterioration in distributional accuracy was directly linked to reduced classifier performance, underscoring the limitations of current LLMs in maintaining data realism at scale. Furthermore, our privacy assessments revealed that while low-dimensional synthetic datasets provided robust privacy preservation against membership inference attacks, higher-dimensional datasets became increasingly vulnerable. Elevated ROC AUC and membership advantage metrics in high-dimensional settings indicated that adversaries could more easily distinguish between members and non-members, thereby compromising patient privacy.

Additionally, our analysis demonstrated that increasing the sample size from 1,000 to 10,000 records consistently improved both the fidelity and privacy metrics of the synthetic data. Larger sample sizes facilitated a better approximation of the real data distribution, resulting in lower KL divergence and enhanced classifier performance.

These findings highlight a critical trade-off between feature dimensionality and the ability of LLMs to generate synthetic EHRs that are both accurate and privacy-preserving. To harness the full potential of LLMs in generating synthetic data, future research must address these challenges by developing more sophisticated generative models capable of capturing high-dimensional dependencies without compromising data quality or privacy.

Some closing remarks indicate that within limited feature spaces, their current limitations in handling high-dimensional data underscore the need for continued advancements in generative modeling techniques. Addressing these challenges is essential to ensure that synthetic healthcare data can reliably support clinical research and innovation while upholding the highest standards of data fidelity and patient privacy.

\paragraph{Limitations}

A significant limitation of this study is its focus on a single prediction task—ICU mortality prediction. However we motivate our choice for selecting this single prediction task as we tried to replicate the experimental protocol  with prior work \citep{johnson2018generalizability}. Regardless, testing generalization in other prediction tasks, such as length of stay prediction or readmission forecasting, may reveal additional insights into the capabilities and shortcomings of LLM-based synthetic data generation. 

Furthermore, this study solely evaluates generalizability using the eICU dataset. Although eICU serves as a robust benchmark for evaluating synthetic data quality, reliance on a single dataset limits the generalizability of our findings. Real-world EHR systems encompass diverse patient populations, care settings, and data distributions that may not be fully captured in eICU. Testing on datasets such as MIMIC-IV or other EHR databases could uncover dataset-specific biases and challenges in synthetic data generation.

\paragraph{Future Work}

Future work should address the outlined limitations by exploring multiple prediction tasks, such as sepsis prediction, disease progression modeling, and intervention effectiveness, to evaluate the broader applicability of synthetic data in clinical domains.

Additionally, assessing the temporal aspects of synthetic data generation is critical. Future research should determine whether generative models can capture temporal patterns in Electronic Health Records, such as trends in laboratory values and vital signs. 

\paragraph{Acknowledgments}

This project was selected as part of the UCLA Call for OpenAI Project Proposals, and received ChatGPT Enterprise licenses as well as in-kind support from the AI Innovation Initiative. YL, ZY and SL were all contributing members related to this initiative.

\bibliography{chil-sample}
\onecolumn
\appendix

\newpage

\section{Prompt Engineering}

Prompt engineering has emerged as a pivotal technique in leveraging the capabilities of large-scale language models (LLMs), enabling them to perform a diverse array of tasks without explicit fine-tuning. In our work, it served as an important part of our generation process and it is worth explaining the fundamental concept behind its innovation. At its core, prompt engineering \citep{giray2023prompt} involves crafting input prompts in a manner that guides the model to generate desired outputs effectively. The theoretical underpinnings of prompt engineering can be elucidated through the lens of information theory and the principles of conditional probability within the framework of probabilistic language models.

Fundamentally, language models like GPT-4 are trained to predict the next token in a sequence, effectively modeling the conditional probability distribution \( P(w_{t} | w_{1}, w_{2}, \ldots, w_{t-1}) \). Prompt engineering manipulates the initial sequence \( w_{1}, w_{2}, \ldots, w_{k} \) to condition the model's predictions towards a specific subspace of the output distribution. The efficacy of a prompt can thus be viewed as its ability to increase the mutual information between the prompt and the desired output, effectively narrowing the entropy of the target distribution in a controlled manner.

One can formalize this by considering the Kullback-Leibler (KL) divergence between the model's output distribution conditioned on the engineered prompt \( P_{\text{prompt}}(w) \) and an idealized target distribution \( P_{\text{target}}(w) \). The objective of prompt engineering can be framed as minimizing \( D_{\text{KL}}(P_{\text{target}} \| P_{\text{prompt}}) \), thereby ensuring that the engineered prompt steers the model's output distribution closer to the desired outcome. This minimization aligns the prompt with the intrinsic representations learned during the model's pre-training phase, exploiting the latent knowledge embedded within the model.

From a theoretical perspective, the success of prompt engineering can also be attributed to the model's capacity to generalize from its training data. The prompt serves as a context that activates relevant pathways in the model's deep neural architecture, effectively retrieving and recombining stored knowledge to address specific tasks. This mechanism can be related to the concept of context-dependent representations in neural networks, where the context provided by the prompt modulates the activation patterns across layers, facilitating task-specific behavior without altering the model's parameters.

\newpage
\subsection{Prompts}

In this section we share the prompts used to generate the EHR. We also reference in each prompt that we attach a portion of the data into the prompt. This can be attached if made available or also passed in as a json if using an API. Giving a few samples (e.g. <100) is enough to give the LLM an idea how to replicate the structure.

\begin{tcolorbox}[colframe=black,colback=white,arc=5mm,title=Prompt for Naive Synthetic EHR Generation,fonttitle=\bfseries]
{\small \textcolor{blue}{\textit{[Annotation: This approach is a straightforward (``naive'') generation of synthetic data 
that attempts to preserve the original dataset’s statistical properties while avoiding direct replication. 
It does not explicitly enforce schema constraints or advanced conditioning strategies.]}}} \\

You are an advanced AI model tasked with generating realistic synthetic Electronic Health Records (EHR) while ensuring privacy and compliance with healthcare regulations. \\

Please analyze the attached file, which contains a structured version of the eICU dataset. Your goal is to generate synthetic patient records that preserve the statistical and structural properties of the original dataset while ensuring no real patient data is replicated.\\

Output the synthetic EHR in a structured format such as CSV following the schema of the provided dataset.

\end{tcolorbox}

\begin{tcolorbox}[colframe=black,colback=white,arc=5mm,title=Prompt for Synthetic EHR Generation (Schema-Based),fonttitle=\bfseries]
{\small \textcolor{blue}{\textit{[Annotation: This schema-based approach enforces explicit adherence to the data types, relationships, and constraints 
defined in the schema. It ensures logical consistency and structural fidelity, 
but does not rely on incremental (conditional) generation or subgroup-specific modeling.]}}} \\

You are an advanced AI model tasked with generating realistic synthetic Electronic Health Records (EHR) while ensuring privacy and compliance with healthcare regulations.\\

Please generate synthetic patient records following the schema provided in the attached file. Ensure that the synthetic data adheres to the same structural and statistical properties as the schema while introducing sufficient variation to maintain realism.\\

Key considerations:
\begin{itemize}
    \item Strictly follow the data types, constraints, and relationships defined in the schema.
    \item Maintain logical consistency between attributes (e.g., diagnoses should align with prescribed medications).
    \item Generate a diverse set of synthetic patient profiles with varying conditions and treatments.
    \item \textcolor{red}{[ADDITIONAL CONSTRAINTS]}
\end{itemize}

Output the synthetic EHR in a structured format such as CSV following the schema of the provided dataset.

\end{tcolorbox}

\begin{tcolorbox}[colframe=black,colback=white,arc=5mm,title=Prompt for Generation of Synthetic EHR (Conditional),fonttitle=\bfseries]
{\small \textcolor{blue}{\textit{[Annotation: This prompt emphasizes conditional generation, in which features are sampled 
sequentially while conditioning on previously generated attributes. It preserves nuanced interdependencies 
(e.g., age-informed vitals) and generates data step by step, aligning with real-world statistical relationships.]}}} \\

You are an advanced AI model designed to generate realistic synthetic Electronic Health Records (EHR) using a conditional generation strategy. Unlike purely schema-constrained approaches, this method incrementally samples each feature while conditioning on previously generated data, ensuring dynamic coherence across patient attributes.\\

\textbf{Generation Process:}
\begin{enumerate}
    \item Start with an initial set of patient attributes from the attached dataset \textcolor{red}{[DATA\_0]}.
    \item Sequentially generate each new feature $x_i$, conditioning on all prior features $(x_1, \dots, x_{i-1})$ to preserve statistical dependencies.
    \item Attach the newly generated data at each step as \textcolor{red}{[DATA\_x\_i]} and proceed iteratively.
    \item Maintain realistic clinical relationships (e.g., heart rate trends consistent with age, plausible lab value correlations, etc.). For example, if $x_1$ (age) = 75, the model should conditionally generate $x_2$ (heart rate) using geriatric norms.
\end{enumerate}

\textbf{Expected Output:}
\begin{itemize}
    \item Output the synthetic EHR in a structured format such as CSV following the schema of the provided dataset.
    \item The sequence of generated features should be iteratively saved in \textcolor{red}{[DATA\_x\_i]} to facilitate stepwise conditioning.
    \item Generated records must align with known medical distributions and avoid contradictions (e.g., incompatible comorbidities).
\end{itemize}

\end{tcolorbox}

\begin{tcolorbox}[colframe=black,colback=white,arc=5mm,title=Prompt for Generation of Synthetic EHR (Group),fonttitle=\bfseries]
{\small \textcolor{blue}{\textit{[Annotation: This approach organizes synthetic data generation by demographic groups (e.g., sex, race). 
It enforces that group-specific distributions and medical patterns are preserved (e.g., female hemoglobin levels), 
ensuring more granular realism within each subgroup.]}}} \\

You are an advanced AI model designed to generate realistic synthetic Electronic Health Records (EHR) using a group-based generation strategy. This method ensures that synthetic records preserve the statistical properties of different demographic groups while maintaining coherence within each subgroup (e.g., SEX, RACE).\\

\textbf{Generation Process:}
\begin{enumerate}
    \item Start with an initial group-defining feature \textcolor{red}{(e.g., SEX or RACE)} from the attached dataset.
    \item Sequentially generate a set of features $x_i$'s, conditioning on the group identity.
    \item Maintain demographic-specific medical distributions, ensuring that:
    \begin{itemize}
        \item Certain conditions/disease risks vary appropriately by group.
        \item Lab values and vitals reflect known variations across demographics.
        \item Medication and treatment patterns align with clinical norms for the given group.
    \end{itemize}
\end{enumerate}

For example, if $G$ = Female and $x_1$ (Age) = 65, then $x_2$ (Hemoglobin Levels) should follow distributions observed in elderly female populations. \\

\textbf{Expected Output:}
\begin{itemize}
    \item Output the synthetic EHR in a structured format such as CSV following the schema of the provided dataset.
    \item Generated records must reflect realistic group-level medical trends while avoiding biases or inconsistencies.
\end{itemize}

\end{tcolorbox}

\newpage
\section{Dataset Characteristics}

\begin{table}[h!]
\begin{tcolorbox}[colframe=black!75, colback=white, title=Summary Statistics of Data]
\centering
\caption{Summary Statistics of the eICU Dataset}
\begin{tabular}{ll}
\toprule
                          Statistic &   Value \\
\midrule
                 Sample Size (Rows) &   88857 \\
       Number of Features (Columns) &      83 \\
Percentage of Positive Death Labels &   8.67\% \\
          Number of Female Patients &   40459 \\
            Number of Male Patients &   48398 \\
   Number of Missing Values (Total) & 1896830 \\
\bottomrule
\end{tabular}
\end{tcolorbox}
\end{table}

\newpage

\section{Additional Analysis}

\begin{figure}[h!]
    \centering
    \includegraphics[width=0.5\linewidth]{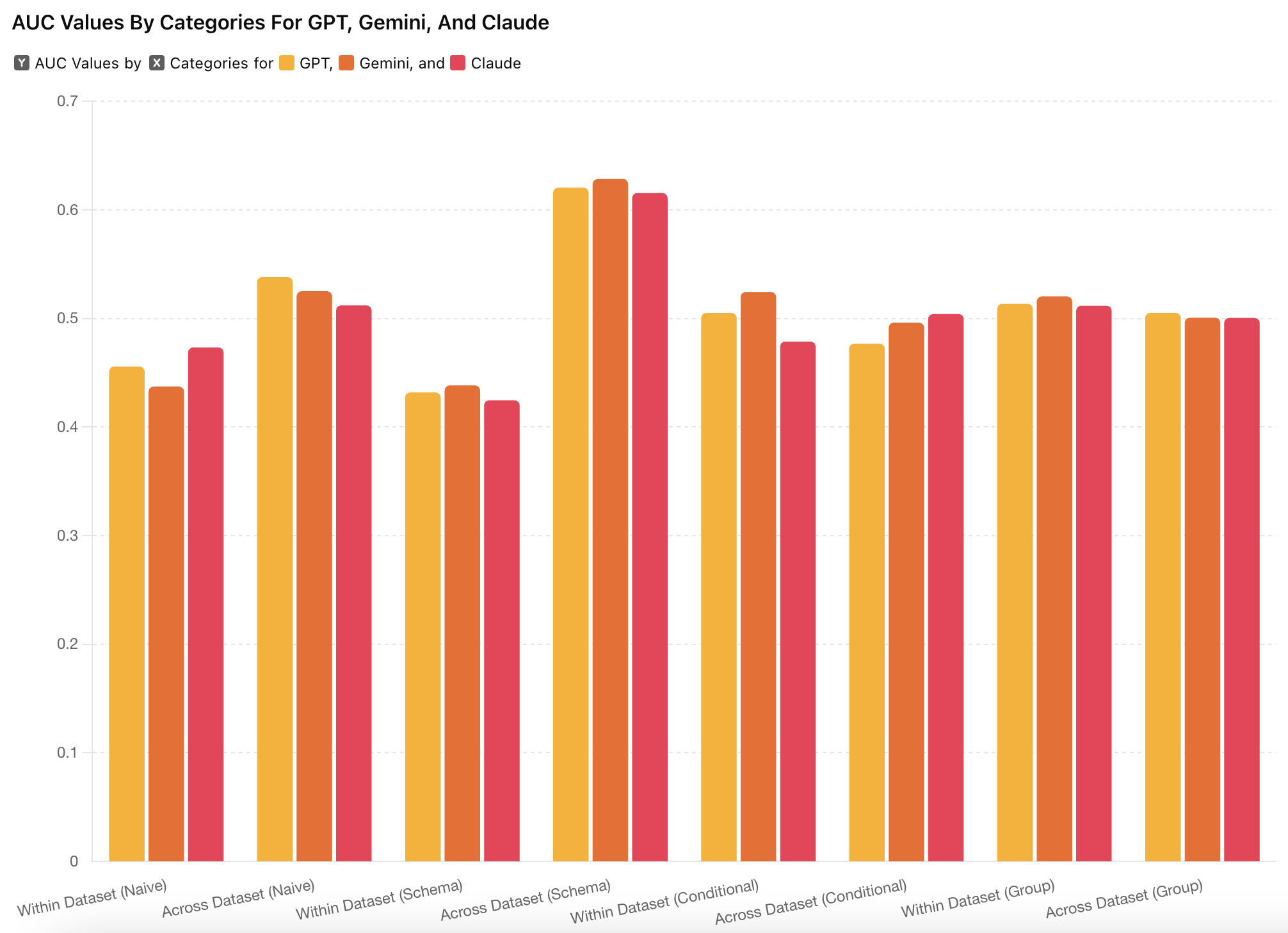}
    \includegraphics[width=0.5\linewidth]{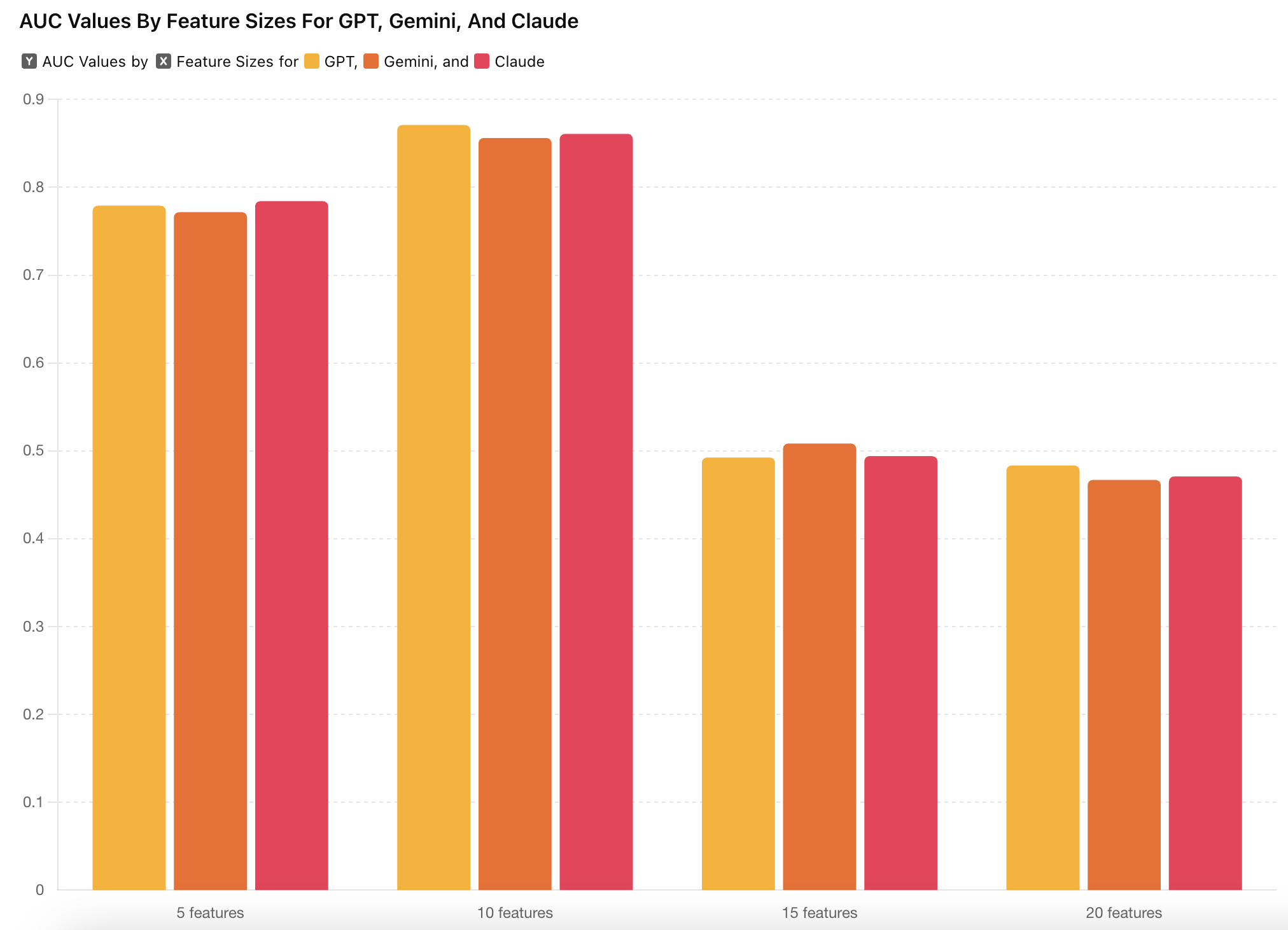}
    \includegraphics[width=0.5\linewidth]{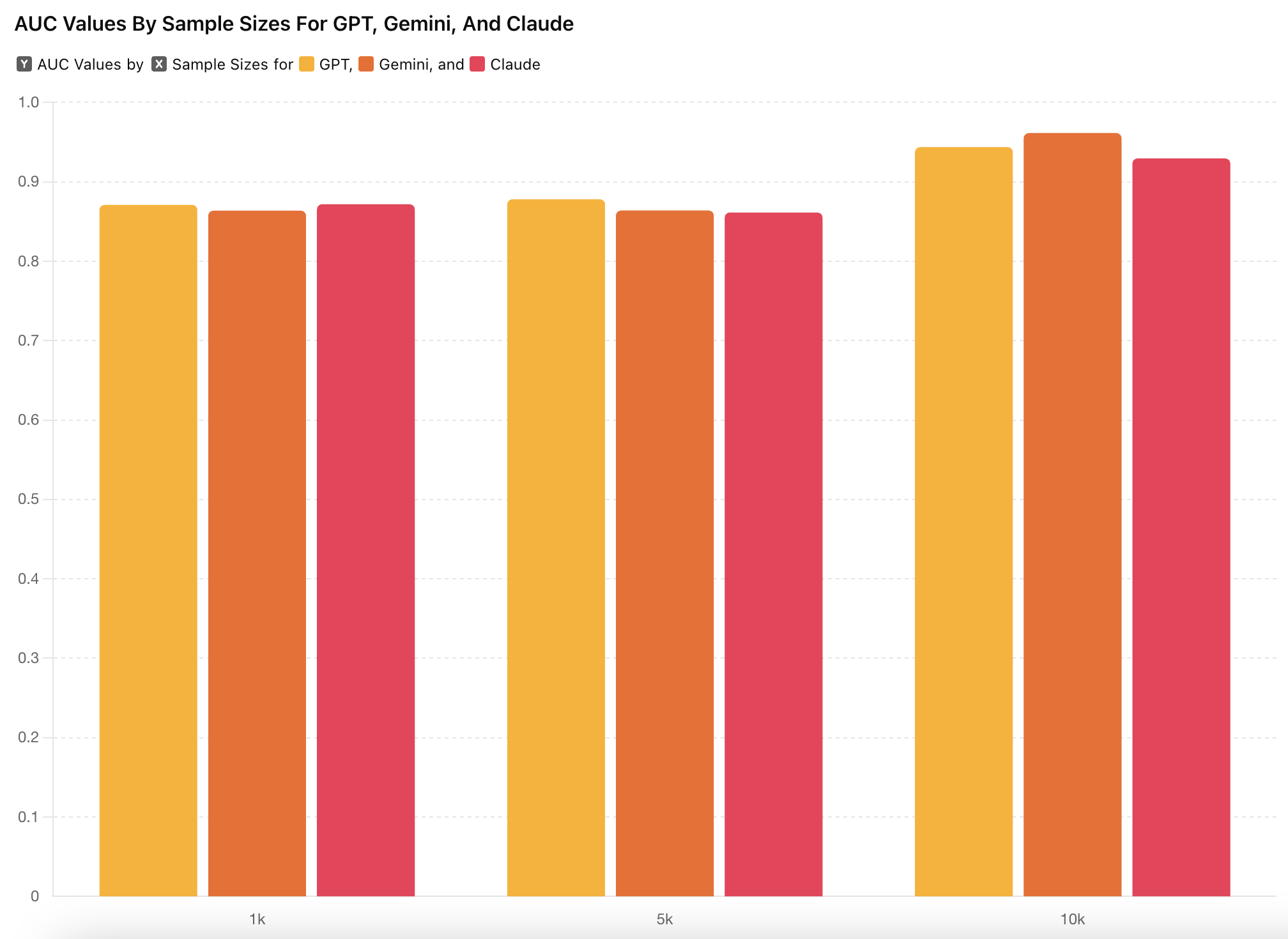}
    \caption{Several bar plots repeating the experiments from the main paper and comparing other commercial LLMs: Gemini and Claude.}
    \label{fig:enter-label}
\end{figure}

In this analysis, we examine and extend the analysis of the main paper and compare the performance of three models—GPT, Gemini, and Claude—under varying conditions of feature sizes, sample sizes, and task categories. The primary metric, Area Under the Curve (AUC), provides a comprehensive measure of the models' ability to distinguish between outcomes (mortality) across these configurations. The results highlight nuanced patterns in model behavior, particularly in relation to data dimensionality and complexity, offering insights into their generalization capabilities.

\noindent \paragraph{ AUC Across Different Generation Strategies}

Similar to the main paper, we tasked each LLM with generating synthetic data using four different generation methods, producing 1,000 samples across all 83 features. The results reveal that all commercial LLMs struggle to generate data at this level of sophistication. As discussed in the main paper, this is likely due to the presence of adverse features and covariates that lack proper relationships, inadvertently affecting the models' ability to predict outcomes effectively. To further investigate, we conducted additional tasks to determine whether all commercial LLMs exhibit similar patterns consistent with our main findings.

\noindent  \paragraph{ Feature Size and the Curse of Dimensionality}

The second dimension of interest examines the models' performance across varying feature sizes. Increasing the number of features introduces higher dimensionality, which poses challenges related to sparsity, overfitting, and increased complexity in learning meaningful patterns. All three models achieve their peak performance with a feature size of 10, suggesting that this is a critical point where the dimensionality provides sufficient information without overwhelming the models' capacity.

\noindent  \paragraph{ Sample Size and Learning in Low- and High-Data Regimes}

The third dimension evaluates how the models respond to varying sample sizes, ranging from small datasets with 1,000 samples to larger datasets with 10,000 samples. As expected, all models benefit from increased sample sizes, with AUC values improving significantly between 1,000 and 5,000 samples and plateauing as the sample size approaches 10,000. This trend underscores the importance of larger datasets in reducing noise and enabling the models to capture underlying patterns effectively.

\noindent \paragraph{ Dimensionality and Its Implications for Generalization}

The interplay between dimensionality and model performance provides key insights into the strengths and limitations of these systems. The feature dimensionality, as reflected in varying feature sizes, highlights a trade-off between information richness and the complexity of high-dimensional spaces. Similarly, the dimensionality introduced by sample size underscores the importance of data quantity in model generalization. However, one can also argue that the comparison of these different commercial LLMs yield results that are nearly statistically indistinguishable from one another reinforcing the findings of \citep{lee2024feet}.

\newpage
\subsection{The Best Configurations for Synthetic Data Generation}

\begin{table}[ht!]
\begin{tcolorbox}[colframe=black!75, colback=white, title=Best configurations]
\centering
\begin{adjustbox}{width=0.5\columnwidth}
\begin{tabular}{@{}ll@{}}
\toprule
\textbf{Configuration Parameter} & \textbf{Optimal Value} \\ \midrule
\textbf{Strategy}                & Schema or Group \\
\textbf{Number of Features}      & 10 \\
\textbf{Sample Size}             & 10k \\
\bottomrule
\end{tabular}
\end{adjustbox}
\caption{Optimal Configuration for Synthetic Data Generation}
\label{tab:best_configuration}
\end{tcolorbox}
\end{table}

\begin{table}[ht!]
\begin{tcolorbox}[colframe=black!75, colback=white, title=Best Performance]
\centering
\begin{adjustbox}{width=0.5\columnwidth}
\begin{tabular}{@{}ll@{}}
\toprule
\multicolumn{2}{c}{\textbf{Performance Metrics (10 Features / 10k Samples)}} \\
\midrule
\quad Avg. KL Divergence (Across) & 0.0821 \\
\quad AUC (Across)                & 0.9157 [0.9089, 0.9217] \\
\quad AUPRC (Across)              & 0.7969 [0.7797, 0.8122] \\
\bottomrule
\end{tabular}
\end{adjustbox}
\caption{Performance values at optimal configuration for Synthetic Data Generation}
\label{tab:best_configuration}
\end{tcolorbox}
\end{table}

\newpage
\section{Additional Figures}

\begin{figure}[h!]
    \centering
    \includegraphics[width=0.42\linewidth]{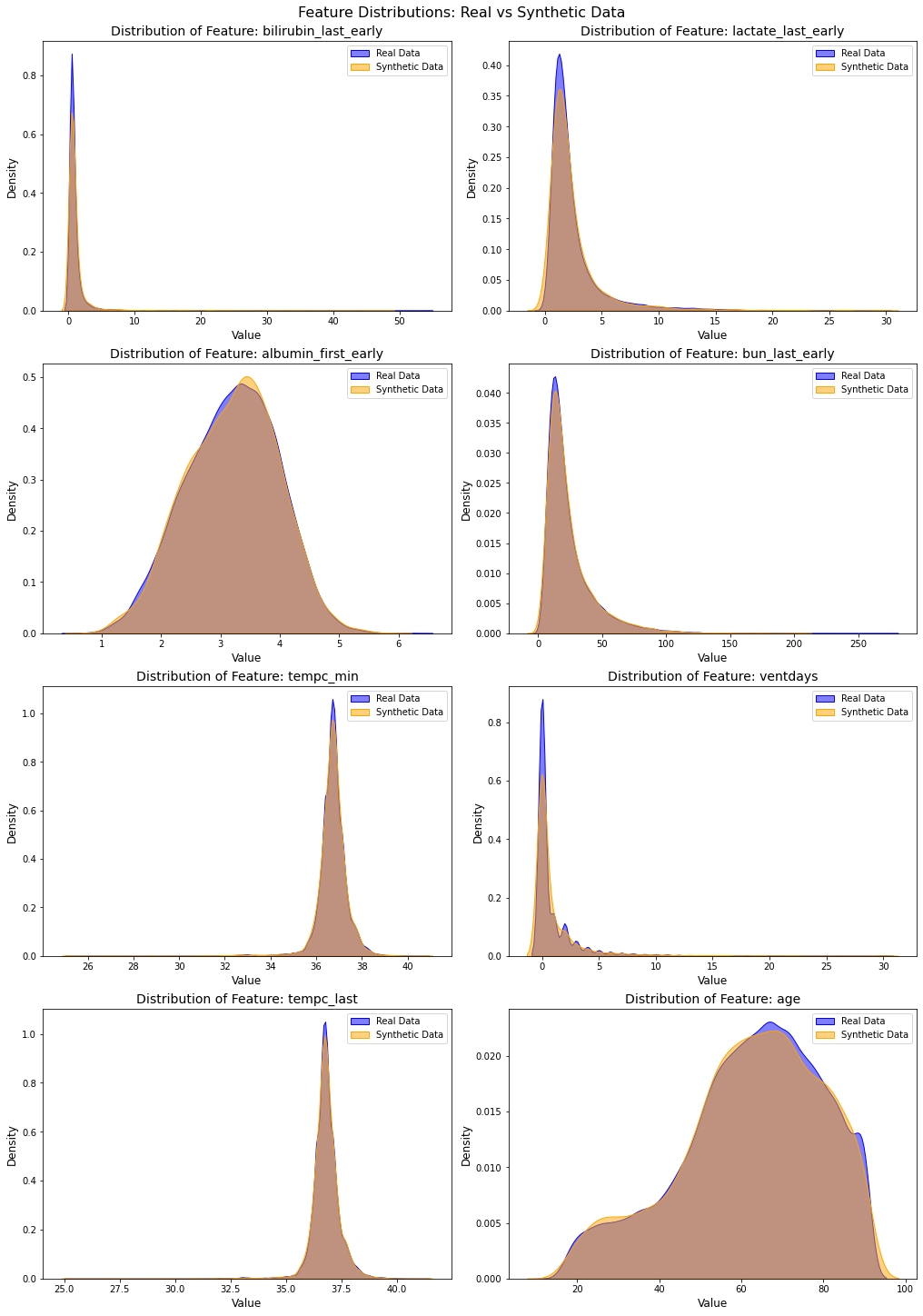}
    \includegraphics[width=0.42\linewidth]{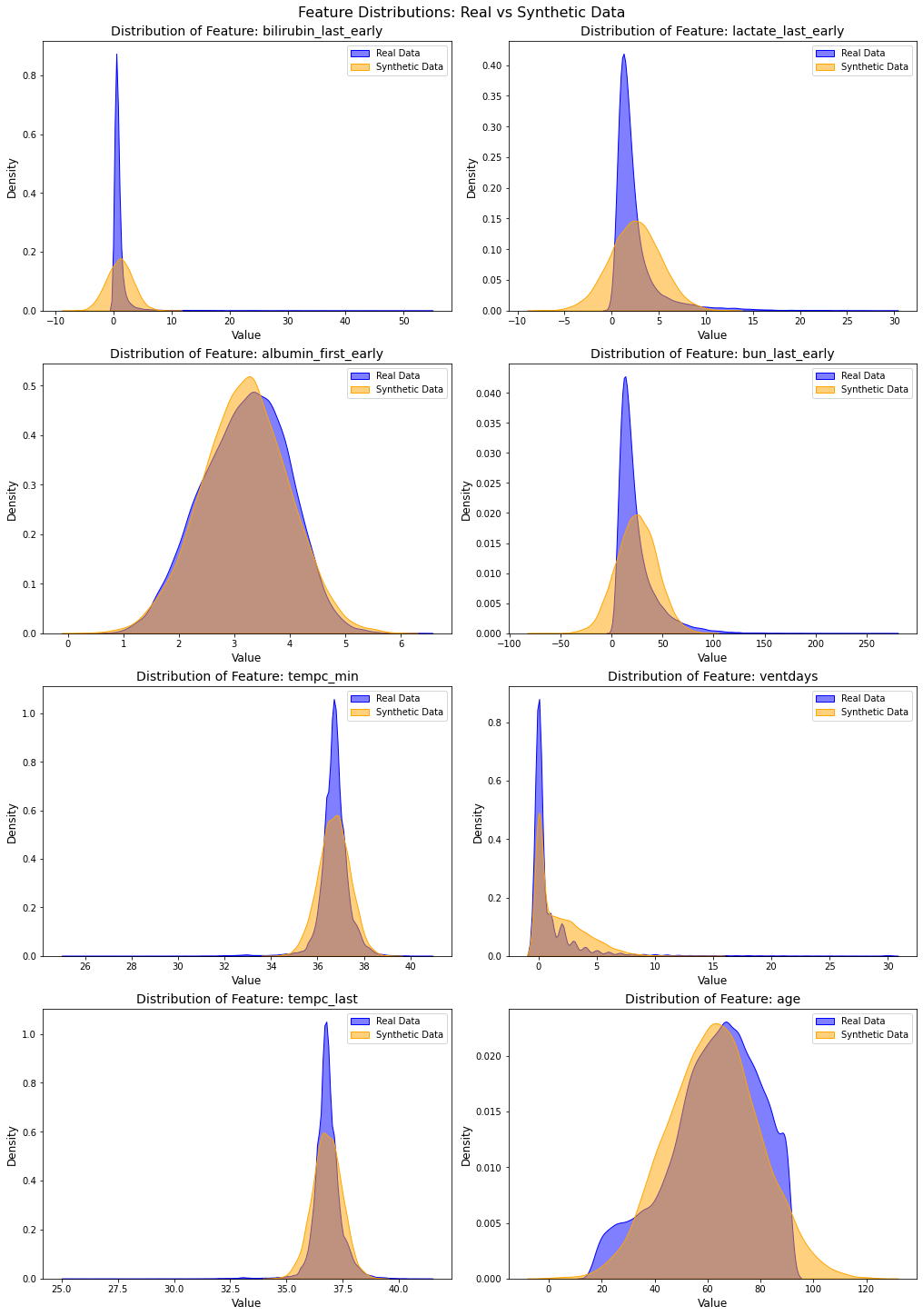}
    \caption{The comparison of distributions between an LLM asked to generate 10 features versus all 83. We only plot continuous features but we see substantial differences in synthetic data generation fidelity.}
    \label{fig:enter-label}
\end{figure}

\begin{figure}[h!]
    \centering
    \includegraphics[width=0.5\linewidth]{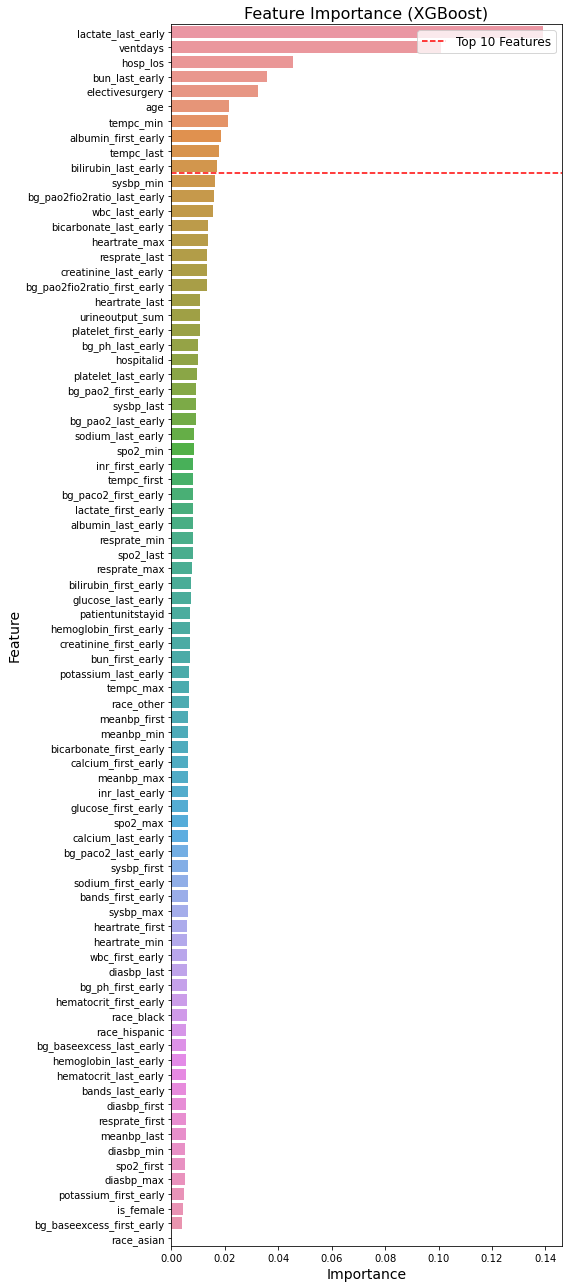}
    \caption{Feature Importance Plot to help with our feature selection study}
    \label{feature-importance}
\end{figure}

\newpage

\section{Metrics}

\subsection*{Area Under the Receiver Operating Characteristic Curve (AUROC)}
\begin{equation}
\text{AUROC} = \frac{1}{n_1 n_0} \sum_{i=1}^{n_1} \sum_{j=1}^{n_0} \mathbb{I}(s_i > s_j)
\end{equation}
where $n_1$ and $n_0$ are the number of positive and negative samples, $s_i$ and $s_j$ are the scores for positive and negative samples, respectively, and $\mathbb{I}(\cdot)$ is the indicator function.

\subsection*{Area Under the Precision-Recall Curve (AUPRC)}
\begin{equation}
\text{AUPRC} = \int_0^1 \text{Precision}(r) \, \text{d}\text{Recall}(r)
\end{equation}
where $\text{Precision}(r)$ and $\text{Recall}(r)$ are the precision and recall at a given threshold $r$.

\subsection*{Kullback-Leibler (KL) Divergence}
\begin{equation}
D_{\text{KL}}(P \parallel Q) = \int_{\mathcal{X}} P(x) \log \frac{P(x)}{Q(x)} \, \mathrm{d}x
\end{equation}

\subsection*{Membership Advantage}
\begin{equation}
\text{Membership Advantage} = \max_s \left| \mathbb{P}(s \mid \text{member}) - \mathbb{P}(s \mid \text{non-member}) \right|
\end{equation}
where $s$ is the score, and $\mathbb{P}(\cdot)$ represents the probability distribution over scores.

\subsection*{Empirical Risk}
\begin{equation}
\mathcal{R}(h) = \frac{1}{n} \sum_{i=1}^n \ell(h(x_i), y_i)
\end{equation}
where $h$ is the hypothesis, $\ell(\cdot, \cdot)$ is the loss function, $x_i$ and $y_i$ are the input and label for the $i$-th sample, and $n$ is the total number of samples.

\end{document}